# The formation of perceptual space in early phonetic acquisition: a cross-linguistic modeling approach

Frank Lihui Tan & Youngah Do (The University of Hong Kong)

{frank.lhtan@connect.hku.hk, youngah@hku.hk}

**Abstract**

This study investigates how learners organize perceptual space in early phonetic acquisition by advancing previous studies in two key aspects. Firstly, it examines the shape of the learned hidden representation as well as its ability to categorize phonetic categories. Secondly, it explores the impact of training models on context-free acoustic information, without involving contextual cues, on phonetic acquisition, closely mimicking the early language learning stage. Using a cross-linguistic modeling approach, autoencoder models are trained on English and Mandarin and evaluated in both native and non-native conditions, following experimental conditions used in infant language perception studies. The results demonstrate that unsupervised bottom-up training on context-free acoustic information leads to comparable learned representations of perceptual space between native and non-native conditions for both English and Mandarin, resembling the early stage of universal listening in infants. These findings provide insights into the organization of perceptual space during early phonetic acquisition and contribute to our understanding of the formation and representation of phonetic categories.

*Keywords*: language acquisition, phonetics, perception, autoencoder, bottom-up learning

## 1  Introduction

Many studies have demonstrated that, during the first six months, infants can distinguish phonetic contrasts not only in their native language but also in non-native languages. However, their ability to perceive sounds in non-native languages gradually diminishes, as they develop better sound discrimination ability tuned to their native language (Aslin et al., 1981; Best et al., 1988; Bosch & Sebastián-Gallés, 2003; Polka & Werke, 1994; Streeter, 1976; Syndral-Lasky &

Klein, 1975; Trehub, 1976; Tsao et al., 2006; Werker et al., 1981; Werker & Lalonde, 1988; Werker & Tees, 1984). For instance, both English and Mandarin native infants exhibit similar abilities to distinguish the Mandarin alveolo-palatal affricate-fricative contrast at the age of 6 months. However, at 12 months of age, English native infants display a decline in their ability to distinguish these sounds, while Mandarin native infants show an enhanced sensitivity (Tsao et al., 2006). As Kuhl (1993) described, infants transition from being "citizens of the world" to becoming "culturally bound listeners".

During the early acquisition phase, infants' cognitive representation of individual sounds is traditionally conceptualized as "phonetic categories", which are established prior to the mastery of phonology (Best, 1994; Kuhl, 1993; Kuhl & Iverson, 1995; Kuhl et al., 1992; Kuhl et al., 2008; Werker & Curtin, 2005). It is hypothesized that phonetic categories are formed through infants' exposure to the distributional properties of acoustic cues (Meye et al., 2002; Werker et al., 2012). Recent studies have additionally suggested that, at this stage, infants adapt their perceptual space to maximize phonetic contrast through distributional learning, without necessarily forming clear-cut phonetic categories (Feldman et al., 2021; McMurray, 2023). The acquisition of the pre-phonological, cross-linguistic perceptual map begins remarkably early in human development. Infants demonstrate sensitivity to vowel contrasts as early as one month of age (Trehub, 1973). Furthermore, between 2-4 months old, infants are capable of categorizing varying instances of the same vowel as one constant sound (Kuhl, 1979).

In this early acquisition stage, infants lack understanding of higher-level linguistic units, such as phrases or sentences. As a result, they can hardly integrate lexical or structural information into the learning of phonetic categories or perceptual map (Locke, Studdert-Kennedy 1983; Boersma 2011; Pierrehumbert 1990; Hayes, 2004). This condition, characterized by a lack of higher-level linguistic information, is referred to as "bottom-up" learning in previous modeling research, as seen in works on the early stages of speech acquisition (Le Calvez & Peperkamp, 2007; Matusevych et al., 2023; Shain & Elsner, 2019). As phonological acquisition progresses, syntactic or semantic information can also be incorporated into phonological knowledge in a "top-down" manner. See Boersma (2011) for a discussion on the specific roles of bottom-up and top-down elements to phonological learning.

Both top-down and bottom-up approaches have been actively studied using computational models. One set of top-down models has been trained on text data, including orthographies, phonemic transcriptions, and phonetic transcriptions (Kolachina & Magyar, 2019; Silfverberg et al., 2018; Sofroniev & Çöltekin, 2018). Other works include word-level cues alongside acoustic inputs (Feldman et al., 2013; Kamper et al., 2015; Theolliere et al., 2015). Despite its efficiency and success, top-down learning models may not accurately reflect empirical processes during the stages of phonetic category learning or perceptual mapping, primarily because infants lack structural knowledge of language during the initial stages of acquisition.

A substantial body of work has been motivated more by empirical evidence, exploring the learning of phonetic categories or perceptual map through the use of bottom-up data. Early attempts tended to rely on manually selected features such as formants (Guenther & Gjaja, 1996; Vallabha et al., 2007) or voice onset time (McMurry et al., 2009; Toscano & McMurray, 2010). Later models sought to learn directly from audio data, instead of abstract acoustic features. For instance, deep convolutional Generative Adversarial Network (GAN) was applied to learn from pure audio data in an unsupervised manner. Beguš (2020b) trained the waveGAN model on continuous speech clips and showed that bottom-up models trained on raw sound input can effectively capture sound features with distinct latent variables and encode phonological knowledge. GAN models were also applied to other phonological learning tasks such as learning of reduplication (Beguš, 2021b), non-local alternation (Beguš, 2022), iterative learning (Beguš, 2020a), articulation acquisition (Beguš et al., 2022), as well as incorporating lexical meaning (Beguš, 2021a). The model parameters were found to be similar with brain stem signals (Beguš et al., 2023).

Autoencoder structure was also applied to learn representations of phones and words from continuous raw audio data. For instance, Eloff et al. (2019) used a CNN-autoencoder with discretized hidden representation to reconstruct acoustic information from fixed-length input. The hidden representation achieved high accuracy in an ABX test. Chen and Hain (2020) and Chorowski et al. (2019) conducted comparable experiments using a CNN structure in the encoder and WaveNet as the decoder, yielding similar success in distinguishing phone categories. Chung et al. (2016) used sequence-to-sequence autoencoder which learned from acoustic data at word-level. It was found that the model is able to learn a hidden representation that had similar properties as word embedding, where words with lower phoneme sequence edit distance had higher cosine similarity between their respective hidden representation vectors.

Schatz et al. (2021) directly compared model learning results with empirical data on infant's phonetic acquisition. The study employed Dirichlet Process Gaussian Mixture Model to learn on continuous audio data and evaluated the models using ABX test for phoneme discrimination. The models, trained on English and Japanese, demonstrated better performance in discriminating phonemes of their respective native language compared to non-native language. In the discrimination of the /ɹ/ and /l/ contrast, which is present in English but absent in Japanese, the model trained on English exhibited significantly lower ABX errors compared to its Japanese counterpart. However, no performance difference was observed in contrasts common to both languages. These findings echo empirical observations among infants; both American and Japanese infants at 6-8 months could discriminate /ɹ/ and /l/, but by 10-12 months, American infants' ability in its discrimination increased, whereas Japanese infants' sensitivity to this contrast declined (Tsushima et al., 1994). Matusevytch et al. (2023) further evaluated different models on different languages following Schatz et al. (2021).

The present study builds upon the existing literature on early acquisition of phonetic categories or perceptual map, but expands this line of research in two aspects, as follows. First, many of the studies reviewed above have evaluated bottom-up models' effectiveness in phonetic learning by checking whether the model can distinguish phone categories. This provides a direct assessment of phone discrimination ability. However, knowing that the model can make such categorical distinctions may not further display how exactly the learned hidden representation is organized. Crucially, it remains unclear whether the organization of representations is based on phones or features. If organized by phones, the learned hidden representations effectively distinguish between different phones or phonemes, e.g., /t/ vs. /p/. However, they may lack knowledge about the relative position or distance between phones; a model may distinguish between phones like /t/, /d/, /p/, and /b/, but their positions in the hidden representation could be randomly dispersed. In contrast, if hidden representations are organized by-features, distribution would show that, for example, /t/ and /d/ are in a relation similar to /p/ and /b/. Second, the input sets used for training the models thus far in the literature include mainly continuous, unlabeled audio data. This setup closely mimics the auditory input infants experience during early phonetic learning. However, such data includes not only the acoustic information of the sounds themselves but also contextual information embedded in the sequences of sounds, such as the understanding of word-initial [s], syllable-final [t], or intervocalic [d]. However, during the very early stages of language acquisition, infants may not have direct access to such contextual information because they have not yet developed phonological or structural knowledge of language. Shain and Elsner (2019), unlike many previous bottom-up studies, trained an autoencoder model on phonetic information without involving contextual information. The study used bottom-up learning models with stacked fully connected and residual layers to automatically extract phonological features. The model was trained on short, discrete audio segments rather than longer recordings. The learned hidden representation showed partial clustering effect with respect to phonological features. It was observed that certain features were not well organized in the learned phonetic representations, which aligns well with the difficulties observed in infant language acquisition.

Building upon earlier studies, particularly the research conducted by Shain and Elsner (2019) and Schatz et al. (2021), the objective of this paper is to investigate the organization of the learned representations (perceptual space) a learner model constructs based on context-free acoustic information. We aim to understand the extent to which context-free bottom-up acoustic information can bring learners to in terms of phonetic acquisition. We adopt a cross-linguistic modeling and evaluation approach, following the methodology in Schatz et al. (2021). Specifically, we trained models using two languages and tested them on both the languages they were trained in (native conditions) and languages they were not (non-native conditions). This methodology mirrors the experimental conditions used with infants to examine native and non-native language perception.

We demonstrate that unsupervised bottom-up training of the model on context-free acoustic information results in no significant difference between native and non-native discrimination performance, which is similar to the early acquisition stage where infants are still "universal listeners". In addition, training on context-free acoustic data allowed the model to tune its perceptual space and form clusters similar to phonetic categories. These prototypical phonetic categories were primarily organized based on phonological features, aligning with the acoustic foundations of these features.

## 2   Methods

The data and the script containing the data preparation, preprocessing and modeling can be found here: https://osf.io/m4t26/?view_only=6190109e1917434e902ab83f7e29a92b.

### 2.1   Dataset

The experiment was conducted with two typologically unrelated languages, English and Mandarin. English and Mandarin differ in several ways regarding their distinctive phonological feature distributions, such as the contrastiveness of voicing and aspiration (Deterding & Nolan, 2007) among many. For instance, English has a wide range of voiced consonants, including nasals, voiced stops, voiced fricatives, and voiced affricates (Giegerich, 1992, pp. 121-124). And voicing is contrastive among stops, fricatives as well as affricates. In contrast, Mandarin has relatively few voiced consonants, limited to nasals, the lateral, and the voiced retroflex fricative /ʐ/ (Dow, 1972, p. 40). Thus, comparing the learning of the two languages helps explore the extent to which learning models can capture the universality of the phonetic system.

#### 2.1.1   *The English dataset*

We employed the Buckeye Speech Corpus, a time-aligned, phonetically labeled American English speech corpus (Pitt et al., 2007). The corpus comprises approximately 300,000 words spoken by 40 speakers from Central Ohio in conversational settings with an interviewer. High-quality recorders were used to make recordings, resulting in a total of approximately 38 hours of speech data sampled at 16000 Hz with 16-bit depth. All 40 speakers' recordings were included in the English dataset and were used in training, validation, and evaluation of the model. The phonetic labels specified in the corpus are more granular than the English phoneme inventory, which differentiates between nasalized vowels, syllabic nasals and laterals, glottal stop, as well as the intervocalic tap /ɾ/.

#### 2.1.2   *The Mandarin dataset*

We employed AISHELL-3, a large-scale Mandarin speech corpus that comprises over 88,000 read utterances and roughly 85 hours of speech data (Shi et al., 2020). The recordings were made using 44100 Hz, 16-bit HI-FI microphones by 218 native Mandarin Chinese speakers from different provinces. Of the 218 speakers, 64 were randomly selected, ensuring that all chosen speakers had recordings with sufficient number of tokens. The number of speakers was chosen to make the total dataset size compatible with that of the English dataset. Since each Mandarin speaker had fewer sentences than English speakers' data in Buckeye Speech Corpus, 5 English speakers' recordings were roughly equivalent in length to those of 8 Mandarin speakers.

As the Mandarin corpus did not provide phonetic time-alignment, the recordings were automatically aligned using a neural aligner *Charsiu* (Zhu et al., 2022). In addition to the Mandarin phoneme inventory, the pinyin transcription generated by *Charsiu* additionally marked the allophones of /i/ following alveolar fricatives and affricates ([ɿ]) as well as retroflex fricatives and affricates ([ʅ]) (Zee & Lee, 2001). Furthermore, the glide medials and nasal codas were combined with the vowel nucleus into a final form (Lin, 2007) and were not aligned separately.

We want to point out the distinct characteristics of the two corpora: the English corpus consists of interview data, whereas the Mandarin corpus comprises read speech. Despite these differences in the nature of data, we selected these two large-scale corpora to ensure comparability in size. We acknowledge that different nature of data may lead to variations in learning outcomes. To address this issue, we perform cross-linguistic checking of learning. This involves testing an English-trained model with Mandarin data and a Mandarin-trained model with English data. For more details, please refer to Section 3.

## 2.2  Data pre-processing

Prior to utilizing the raw sound data and time-aligned phonetic transcriptions to train and evaluate the computational model, pre-processing procedures were carried out. These procedures included primary feature extraction, segmentation, and chunking. Features were extracted automatically through Python scripts.

### 2.2.1  *Prior extraction of sound data*

We extracted the mel-frequency cepstral coefficients (MFCC) from the raw sound data in advance (Xu et al., 2004). The raw sound data underwent pre-processing by MFCCs using a window length of 25ms and a window step of 10ms, following the default setting as described by Lyons (2013). This process yielded 13 coefficients for each frame, which were generated by extracting wave information in a single window. In addition to the MFCCs, we also extracted differential coefficients (delta) by subtracting the MFCCs of adjacent frames and acceleration (second-order delta) by subtracting adjacent deltas. This allowed us to capture the dynamic trajectories of the MFCCs (Lyons, 2013). The deltas and second-order deltas were concatenated to the MFCCs to create a frame of 39 coefficients.

### 2.2.2   *Sound data segmentation*

To ensure that our experiment focuses solely on the bottom-up learning of phonetic systems, we did not segment the training data according to the phonetic transcriptions. Providing the model with such information would introduce existing boundaries and exclude sounds outside of the segments, which could potentially bias the model. Moreover, there is limited evidence suggesting the existence of primitive sub-word boundaries prior to language acquisition (Kuhl, 2004). Speech input to infants does not appear to provide cues for natural segmental boundaries either (Räsänen, 2014). We used random sampling to extract sound data, resulting in segments with varying lengths that were similar in distribution to the evaluation data with variable segment length. Segments of random lengths were sampled from a truncated normal sampler centered at the mean segment length of the total dataset and lower bounded by 0. This process avoids using fixed-length windows as input and accounts for the inherent variability in segment lengths in natural language. Nevertheless, it must be acknowledged that the process of random segmentation is not entirely devoid of contextual information. This is because both the mean segment length and variance were drawn from the dataset, and therefore the distribution of randomly segmented segments still reflected the language-specific segment length characteristics. The data for evaluation were segmented based on time-aligned phonetic transcriptions, and only segments containing phonetic tokens were included, excluding silence and pure noise.

To simplify modelling and improve accuracy of representation learning, we resampled the segments, which consisted of varying numbers of extracted frames, to a fixed length of 25 frames per segment for input into the autoencoder. This resulted in segment dimensions of $F \times C_i$ ($F = 25$ and $C_i = 39$) for input and $F \times C_o$ ($C_o = 13$) for output. Although the input and output of an autoencoder model are typically identical, our model did not follow this setting because the deltas and second-order deltas were directly derived from the MFCCs.

### 2.3   **Model settings**

The model used in this experiment was an autoencoder (Bank et al., 2020), whose schematic structure is illustrated in Figure 1. As an autoencoder aims to encode input information (acoustic cues) into a latent space (mental representation) and reconstruct the input with minimal distortion, the resulting latent representation can be considered as the learned knowledge of individual segments. As in Figure 1, the model's encoder and decoder each consist of two fully connected layers (FC layer henceforth) and a residual block as the main structure, which was adopted from Shain and Elsner (2019)'s model. The model structure and hyperparameters were pre-tested and tuned to best perform in the segment classification task. The pre-processed input data in the shape of $F \times C_i$ were first flattened to a vector of length $FC_i = 975$ before it was fed into the encoder. This step did not distort the information per se but simply allowed the FC layer to process the short segment as a holistic component, instead of a sequence of frames[1]. After flattening the input data, the encoder transformed the input from $FC$ to an intermediate number of dimension $I = 256$ through a FC layer with ReLU as the activation function following the process in Figure 1.

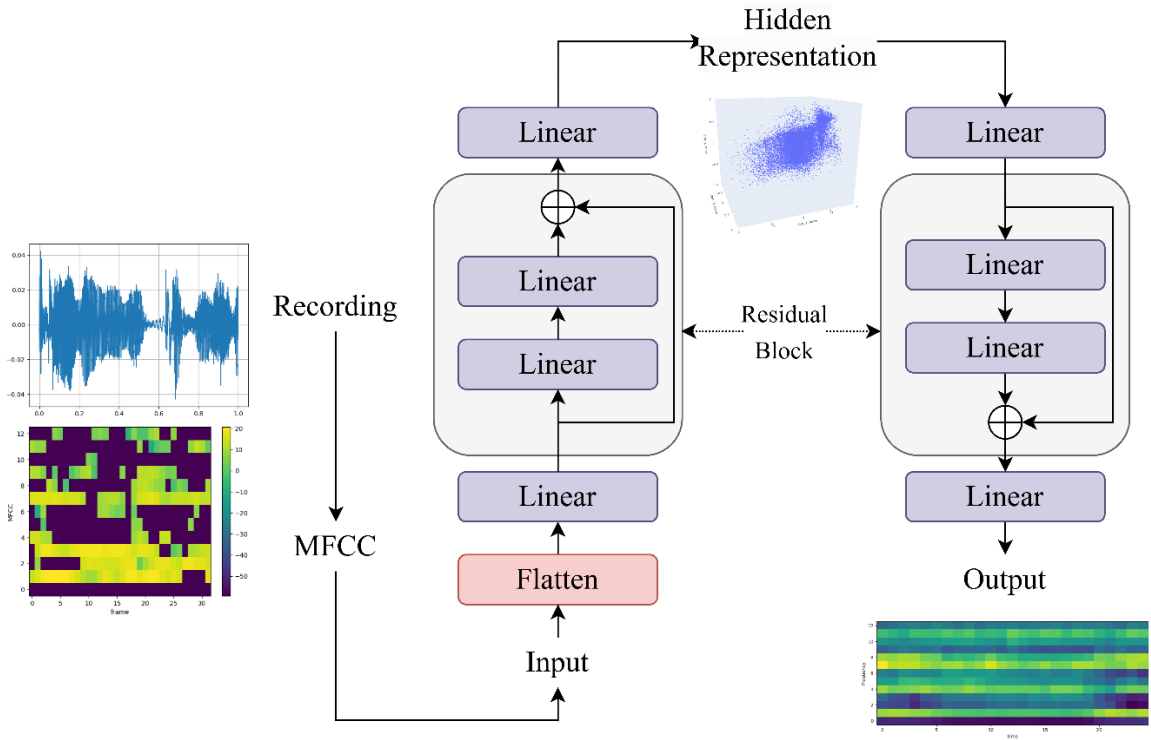

**Figure 1**. Structure of the autoencoder model used for this experiment.

---

[1] Alternative methods might have been Convolutional Neural Network (CNN; Palanisamy et al., 2020) or Recurrent Neural Network (RNN; Sherstinsky, 2020), which can treat input as sequences.

The model structure in Figure 1 is capable of approximating non-linear functions (Liu & Liang, 2019) and is thus able to simulate complex neural activities involved in the speech perception process. The 256-dimensional latent representation was then passed through a residual block consisting of two stacked FC layers with ReLU as the activation function. The residual network structure outperforms the plain linear feedforward network structure in deeper networks due to the shortcuts it provides to propagate information through multiple "paths" with varying network depth (He et al., 2016). A final FC layer transformed the 256-dimensional latent representation into the hidden representation of dimension $H = 3$. The deep neural network model of two FC layers and a residual block (equivalent to a total of four layers) is anticipated to simulate the complex, layered neural transformations underlying speech perception that convert the signal captured by sensory receptors into the underlying neural code of segments (DeWitt & Rauschecker, 2012; Eggermont, 2001; Friederici, 2011; Hickock & Poeppel, 2007; Kell et al., 2018; Yost, 2007).

In our model, the hidden representation was constructed using continuous real number values (set to 3). This differs from the approach of Shain and Elsner (2019), who assumed a hidden representation composed of 8 binary bits. Work on categorical perception (Liberman et al.,1957) suggests that humans perceive sounds as distinct categories, aligning with the assumption made in binary bits in Shain and Elsner (2019). However, human learners go through earlier stages in acquisition where the representations may not be categorical yet (Feldman et al., 2021; McMurray, 2023). To allow this possibility, our approach adopted the continuous representations, mimicking very early stages of acquisition. During later stages of learning, this process may enable the emergence of perceptual categories, as demonstrated in the study by Shain and Elsner (2019).

The model was implemented using Pytorch (Paszke et al., 2017) and was optimized using Adam (Kingma & Ba, 2014) with a learning rate of 0.001. The autoencoder models were trained on 25 English speakers' recordings and 40 Mandarin speakers' recording each, which underwent random sampling and were validated on validation set. The model was trained for 100 epochs and was confirmed to have converged.

## 3  Experimentation and results

To investigate the structure of the hidden representation learned by a model that was primarily trained on bottom-up data and to determine whether the model acquired cross-linguistic knowledge, we carried out the following four simulations: trained in English and evaluated with English (EE), trained in English but evaluated with Mandarin (EM), trained in Mandarin and evaluated with Mandarin (MM), and finally trained in Mandarin but evaluated in English (ME), as outlined in Table 1.

| *Training* <br> *Evaluation* | *English* | *Mandarin* |
|---|---|---|
| *English* | EE (exposed) | ME (foreign) |
| *Mandarin* | EM (foreign) | MM (exposed) |

**Table 1** Four conditions in experiment: training with two languages, each evaluated with the exposed language and foreign language.

Prior to the experiment, the model's structure and hyperparameters were fine-tuned based on validation losses in pre-test as well as the model's clustering performance on the validation set. After that, the experimental pipeline unfolded as follows: first, two identical models were trained—one with English data and another with Mandarin data. Subsequently, clustering test was applied to the evaluation set, and the clustering quality was assessed with homogeneity, completeness, and V-measure (Rosenberg & Hirschberg, 2007). Second, we fed the model with another (foreign) language. For example, the model trained with English data was assessed in clustering evaluation with Mandarin data. The results are in Section 3.1. Higher V-measure score indicates more correct clustering and implies better quality of model's hidden representations. In Section 3.2, we calculated ABX error rates across all phone pairs as well as for specific phonetic contrasts. If the model successfully learned a specific contrast, the error rates from the ABX test will be lower compared to the contrast that has not been acquired. In Section 3.3, the distribution of the learned hidden representations were observed. This illustration represents the specific space that the learner model constructed from the bottom-up learning. Given the large number of phonemes, contrastive features, and phoneme pairs involved, we have only presented representative cases throughout Section 3. However, we have provided the full data sets along with the analysis script in the following link:https://drive.google.com/drive/folders/1ZBjlo2_227Qjh4Oznm-Dgt5WeXi4L1FM?usp=sharing.

## 3.1 Clustering evaluation and results

An unsupervised clustering evaluation using homogeneity, completeness, and V-measure (HCV measures) was performed for all four conditions in Table 1, i.e., for the exposed languages conditions EE and MM, and for the foreign language conditions, i.e., EM and ME. We evaluated the learner model's ability to accurately project different phones to distinct regions of the hidden representation space. The time-aligned evaluation tokens were first projected to the hidden space using the encoder of the trained model. Subsequently, the encoded hidden representations were clustered into 256 clusters using Kmeans, an efficient unsupervised clustering algorithm (Lloyd, 1982). The algorithm partitions data into designated number of clusters based on similarity, and iteratively refines cluster assignments by minimizing the total distances between data points and their cluster centroids. The results were then compared with the ground truth phonetic transcription labels to compute the HCV scores, as displayed in Table 2. The baseline was obtained by randomly assigning labels from the 256 categories to each item in the evaluation set.

| *Model* | *English Evaluation Set* | | | *Mandarin Evaluation Set* | | |
|---|---|---|---|---|---|---|
| | H | C | V | H | C | V |
| *Baseline* | 0.009 | 0.006 | 0.007 | 0.011 | 0.007 | 0.008 |
| *Exposed* | 0.329 | 0.230 | 0.271 | 0.447 | 0.281 | 0.345 |
| *Foreign* | 0.315 | 0.215 | 0.255 | 0.361 | 0.234 | 0.284 |

**Table 2**. Phone clustering scores. Baseline scores indicate the random assignment results on the evaluation set. Foreign language results should be compared to their own baseline instead of that of the exposed language, thus organized according to evaluation language.

As shown in Table 2, the HCV scores of all four conditions, i.e., EE, EM, MM, and ME, were much higher than the baseline model. This observation provides initial support that phonetic category centers emerged from acoustic input that lacked contextual information, e.g., segmental boundary and higher-level structural information, during the autoencoder's learning of reconstructing or reproducing the raw sound input. It can be also observed that when comparing exposed and foreign conditions, the exposed conditions generally achieved higher HCV scores than the foreign conditions, for both English and for Mandarin. This may suggest that the model has developed preferences for the exposed language at this stage to a certain degree.

## 3.2 ABX tests

To investigate whether the model acquired specific phone categories, we conducted the ABX test. Similar to its application in behavioral studies, the error rates from the ABX test indicate which conditions resulted in successful distinction of the sounds by the model, and which conditions led to frequent confusion. We utilized the ABX error rate calculation method described in Schatz (2016) to access the models' phone discrimination performance. We excluded the top and bottom 0.5% of data points in the hidden representations to mitigate the influence of extreme outliers. The hidden representations then underwent z-score normalization. Euclidean distance served as the metric for determining whether a given hidden representation was incorrectly placed closer to another phonetic unit in the calculation of ABX error scores.

We first performed a ABX test encompassing all logically possible vowel and consonant pairs[2]. This generated 885 pairs for the English dataset and 714 pairs for the Mandarin dataset. Overall, the models tested on exposed languages (EE and MM) achieved significantly lower error rate than models tested on untrained, foreign languages (EM and ME) ($\mu_{EE} = 0.3, \sigma_{EE} = 0.12; \mu_{MM} = 0.301, \sigma_{MM} = 0.132; \mu_{ME} = 0.321, \sigma_{ME} = 0.119; \mu_{EM} = 0.327, \sigma_{EM} = 0.126;$ $EE\ vs\ ME, p < 0.001;\ MM\ vs\ EM, p < 0.001;\ EE\ vs\ MM, p = 0.851$). See Figure 2. When combining EE and MM to exposed condition and ME and EM to foreign condition, results also show that on exposed languages the models performed in general significantly better than on foreign languages ($\mu_{Exposed} = 0.3, \sigma_{Exposed} = 0.125; \mu_{Foreign} = 0.324, \sigma_{Foreign} = 0.122;\ p < 0.001$). The result from this AXB test is in line with the higher HCV scores obtained from the exposed languages compared to the foreign languages, presented in Table 2. These results indicate that the models acquired the phone knowledge based on their input and shows preference towards the exposed language (Kuhl et al., 1992). Based on this initial observation, we question whether the models have fine-tuned their perceptual map to native language phones, causing them to lose sensitivity to featural contrasts not present in their native language. Such loss of sensitivity has been widely documented in experiments with infants, as discussed in the literature review in Section 1.

---

[2] We tested all phone category pairs against vowel pairs and consonants pairs. The all-phone category pairs yielded a lower general ABX error rate; however, due to the focus of our study not being on the discrimination between consonants and vowels, we limited our testing to vowel pairs and consonant pairs separately.

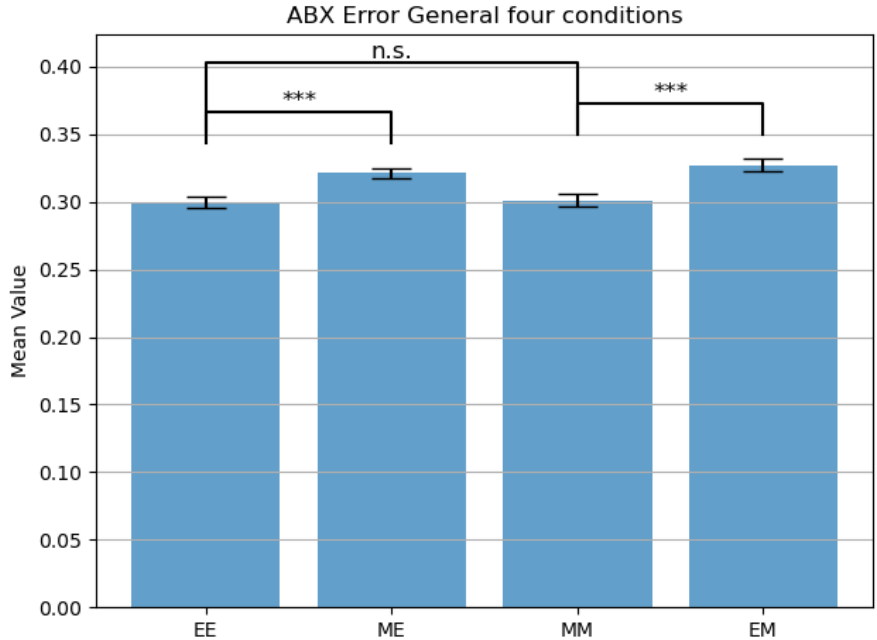

**Figure 2** ABX error in general test for four conditions.

Due to the large number of contrasts, this section selectively presents representative cases of minimally contrastive feature pairs that represent various distributional conditions. For comprehensive data, please refer to the OSF link (p.5). First, we will examine the learning of contrasts that exists only in one language. Second, we will focus on contrasts that are more prevalent in one language compared to the other. Lastly, we will illustrate a case where the contrast is equally common across the two languages. By examining these representative cases, we can better understand how the model acquires and distinguishes different types of contrasts.

First, /ɕ/ – /tɕʰ/ contrast exists in Mandarin but not in English. The ABX test on the /ɕ/ – /tɕʰ/ contrast revealed no significant performance difference between models trained on Mandarin and those trained on English ($\mu_{Exposed} = 0.491, \sigma_{Exposed} = 0.014; \mu_{Foreign} = 0.497, \sigma_{Foreign} = 0.007; p = 0.22$). See Figure 3. Similarly, when tested on the tense-lax vowel contrast in English, which is not contrastive in Mandarin (Chen, 2006), exposed and foreign models displayed no significant difference in ABX test performance ($\mu_{Exposed} = 0.419, \sigma_{Exposed} = 0.044; \mu_{Foreign} = 0.414, \sigma_{Foreign} = 0.048; p = 0.619$). See Figure 4.

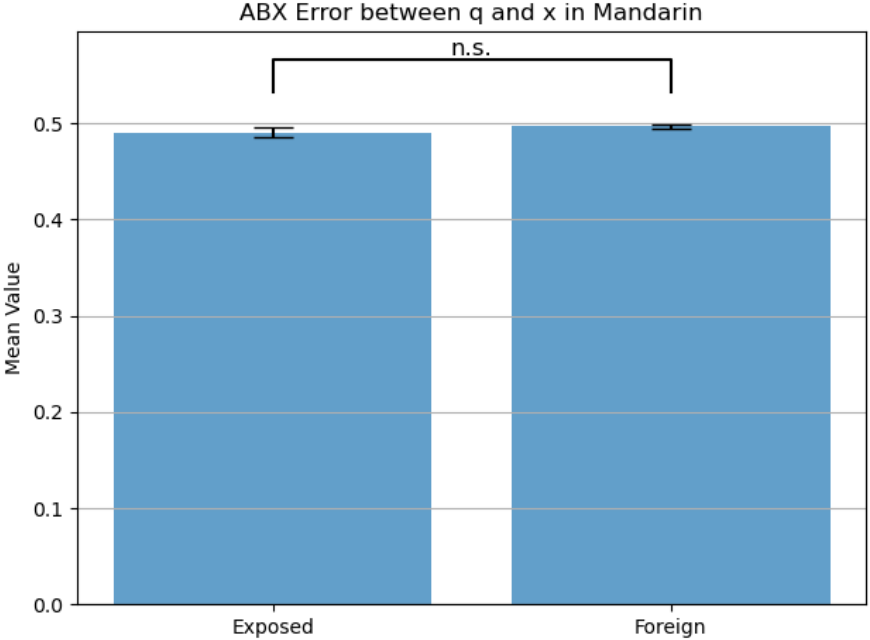

**Figure 3** ABX error on /ɕ/ – /tɕʰ/ contrast for exposed (Mandarin) and foreign (English) language.

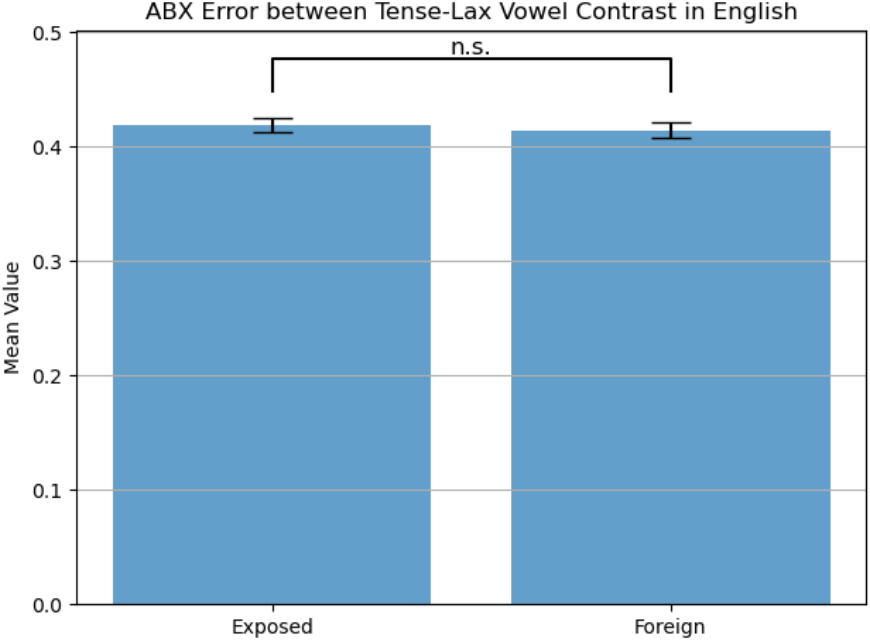

**Figure 4** ABX error on English tense-lax vowel contrast for exposed (English) and foreign (Mandarin) language.

Second, voicing contrast exists in both languages, but it is more prevalent in English than in Mandarin, in terms of relative token frequencies among consonants (English: 30.27%, Mandarin: 7.45%). Specifically, in English, voicing contrast can be found between /p, t, k, f, s, ʃ, tʃ, θ/ and /b, d, g, v, z, ʒ, dʒ, ð/ while in Mandarin it is only found between /ʂ/ and /ʐ/ (Dow, 1972, p. 40). Despite its higher proportions in English than in Mandarin, the results indicated no significant differences in the learning on voicing contrast ($\mu_{Exposed} = 0.367, \sigma_{Exposed} = 0.062; \mu_{Foreign} = 0.375, \sigma_{Foreign} = 0.077; p = 0.695$) under both exposed and foreign conditions. See Figure 5. In contrast, the fricative and affricate contrasts are more prevalent in Mandarin than in English (English: 2.03%, Mandarin: 29.37%), but no significant differences were found between the different language training conditions either, as in Figure 6 ($\mu_{EE} = 0.481, \sigma_{EE} = 0.024; \mu_{MM} = 0.458, \sigma_{MM} = 0.053; \mu_{ME} = 0.475, \sigma_{ME} = 0.027; \mu_{EM} = 0.444, \sigma_{EM} = 0.053; EE\ vs\ ME, p = 0.416; MM\ vs\ EM, p = 0.24; EE\ vs\ MM, p = 0.005$).

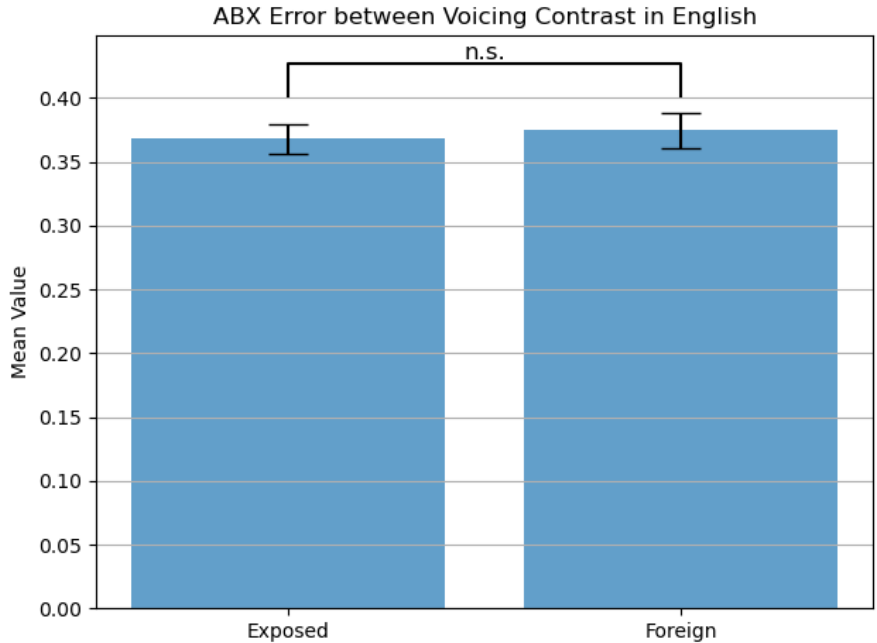

**Figure 5** ABX error on English consonant voicing contrast for exposed (English) and foreign (Mandarin) language.

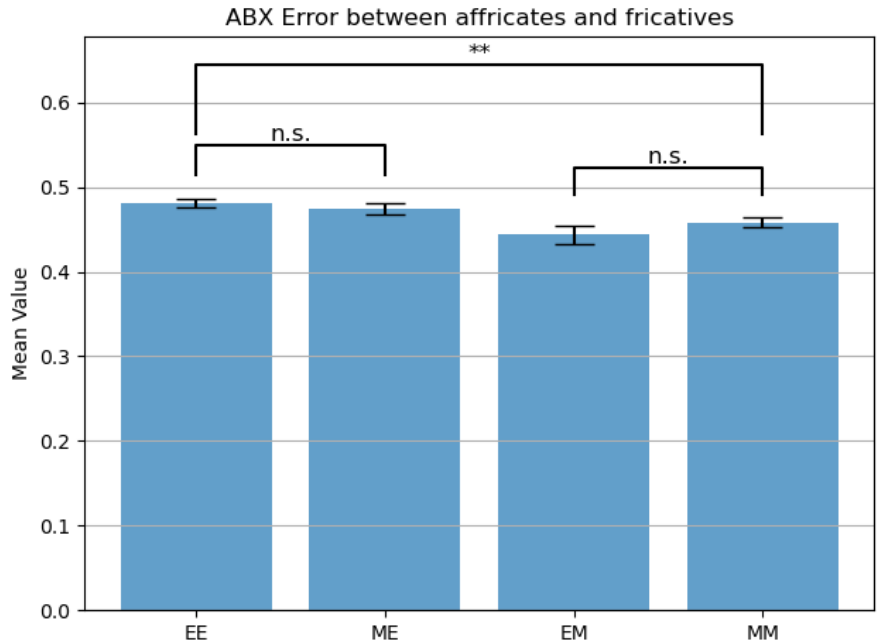

**Figure 6** ABX error on contrast between affricates and fricatives in Mandarin for exposed (Mandarin) and foreign (English) language.

Third, POA contrast among plosives and fricatives is equally common in both languages (plosive: English: 14.73%, Mandarin: 18.71%; fricative: English: 8.08%, Mandarin: 13.75%). As in Figures 7 and 8, the results showed no significant difference between exposed and foreign condition for both languages (plosives: $\mu_{EE} = 0.445, \sigma_{EE} = 0.029; \mu_{MM} = 0.473, \sigma_{MM} = 0.024; \mu_{ME} = 0.455, \sigma_{ME} = 0.03; \mu_{EM} = 0.462, \sigma_{EM} = 0.029; EE\ vs\ ME, p = 0.169; MM\ vs\ EM, p = 0.078; EE\ vs\ MM, p < 0.001$; fricatives: $\mu_{EE} = 0.285, \sigma_{EE} = 0.066; \mu_{MM} = 0.31, \sigma_{MM} = 0.077; \mu_{ME} = 0.35, \sigma_{ME} = 0.096; \mu_{EM} = 0.353, \sigma_{EM} = 0.06; EE\ vs\ ME, p = 0.159; MM\ vs\ EM, p = 0.881; EE\ vs\ MM, p < 0.001$).

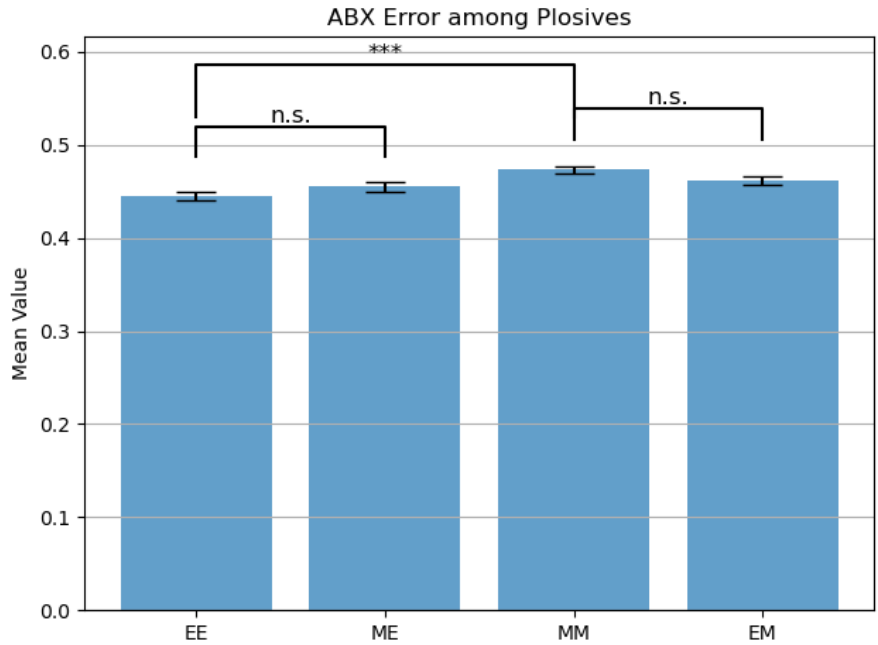

**Figure 7** ABX error on place of articulation contrast among plosives for exposed (Mandarin) and foreign (English) language.

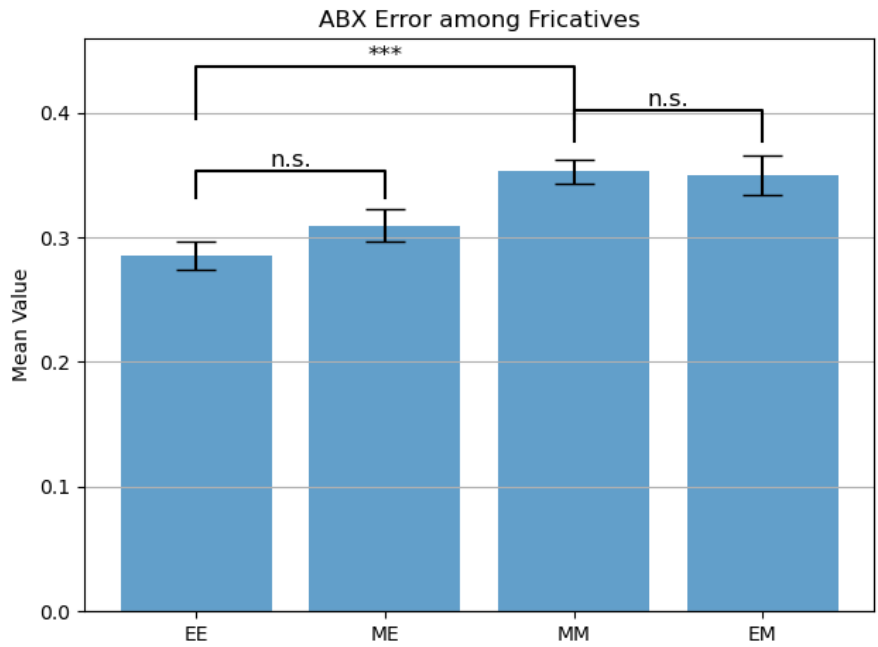

**Figure 8** ABX error place of articulation contrast among fricatives on English for exposed (Mandarin) and foreign (English) language.

The results of the ABX tests on minimally contrastive features suggest that the model's learning does not directly reflect the proportions of the contrasts existent in the language input. Instead, the learner model still maintains universal properties and has not specifically adapted its learned space to language-specific situations. Recall that the models' overall clustering ability (Section 3.1) and the general ABX test results (first part of Section 3.2) indicates that it has overall acquired the input languages, as the model showed higher performance in exposed languages compared to foreign languages. However, when comparing the performances across various phonetic contrasts with different distribution patterns in the input languages (Section 3.2), the results consistently suggest that the model has retained universal phonetic properties and have not yet fine-tuned their perceptual space to their respective native languages on these contrasts. At a first glance, the results from the general ABX test and the specific contrast learning test may appear contradictory. However, we believe this is an expected outcome. The general ABX test includes all possible contrast pairs, including those that are phonologically distant, not in a natural class, and differ in multiple phonological features. Thus, it is reasonable to conclude that the model's higher performance in the general ABX test is a result of successfully learning clearly distinctive contrasts. However, the model is progressing towards acquiring language-specific phoneme inventories but has not yet fine-tuned its learning to distinguish minimal phonetic contrasts that involve smaller phonological distances, as demonstrated in this section. Therefore, the model can still be considered a universal learner in this regard.

## 3.3 Hidden representation

### 3.3.1 *Hidden dimension selection*

Now that we have examined the models' clustering performance (Section 3.1) and categorical phone discrimination performance (Section 3.2), let us proceed to identify the shape of the hidden representations that the models have constructed through learning. To do so, we first conducted a dimensionality search ranging from hidden dimensions of 1 to 15. For each hidden dimension value, the model was trained for 50 epochs, and the epoch with lowest validation loss was utilized to encode the golden validation set segmented with aligned ground truth labels. The encoded hidden representations were clustered and evaluated. As anticipated, the model's efficacy and language learning capabilities increased with a larger hidden dimension. The results plot as shown in Figure 9 illustrates a rather consistent, monotonic rise in performance. Notably, the initial three values of dimensionality exhibited the fastest increase in V-measure scores, with a gradual slowing thereafter. Since our objective is not necessarily to identify the optimal model for the task but rather to examine the similarity of the model's learning process to that of humans, the selection of the hidden dimension was not limited by the requirement for optimal performance. Instead, we chose a dimensionality of 3 for the hidden representation, where the steep increase in V-measure score ceases. We also conducted experiments with larger dimensionalities for the hidden representation. However, the overall pattern of learning results of the model remained consistent regardless of the increase in dimensionality. See the script and the data in the OSF link (p.5).

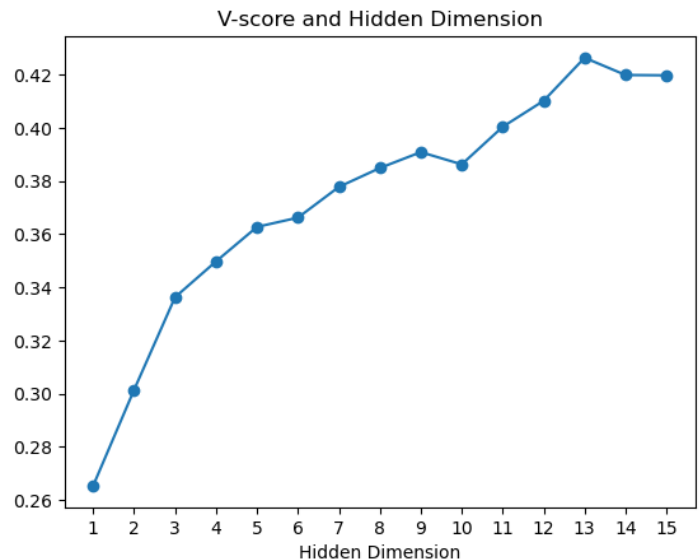

**Figure 9**. The relation between dimensionality of encoder output and model's performance in clustering evaluation (V-measure).

### 3.3.2  *Natural class evaluations*

To assess the model's ability to learn different features and natural classes, we conducted direct plotting of hidden representations. To reflect the gradual binary fission process of human learning (Jakobson & Halle, 1956, p. 47) and the binary nature of distinctive features, pairwise comparisons were performed rather than comparing multiple phonemes together. This approach is supported by evidence that phoneme knowledge in humans emerges from distribution-based clustering involving a series of binary decisions (Guenther & Gjaja, 1996; Hayes, 2004; Maye & Gerken, 2000). We generated natural class pairs that exhibit contrast mostly only in one feature, with a relatively balanced number of members in each class. Please see the OSF link for the data of all logically possible phoneme pairs (p.5).

Prior to conducting the evaluation, similar to the ABX test, we employed the model's encoder to encode the entire evaluation dataset into hidden representations for all simulations, i.e., EE, MM, EM, and ME. Subsequently, the data underwent normalization to facilitate the calculations of silhouette score and hit rate, and to enhance the efficacy of plotting and inter-group comparisons. The top and bottom 0.5% of the data were also excluded in avoidance of extreme outliers especially in foreign language conditions. Following this, min-max normalization was applied, utilizing the minimum and maximum values after filtering for each dimension separately ($\hat{h}_d = 2\frac{h_d - min}{max - min} - 1$). To minimize frequency asymmetry of different phonemes in a language, only around 1500 to 3,000 tokens of each selected phoneme were randomly sampled from the time-aligned evaluation dataset, depending on the available data size. Each token was accompanied by its ground truth label for evaluation. The hidden representations of selected tokens were plotted in three-dimensional scatter plots, where the position of a point in the three-dimensional space indicates the three hidden representation values and color indicates the ground truth label that specifies the phoneme or natural class to which the token belongs based on the ground truth labels. For plotting on two-dimensional media, we selected an angle that provides clear boundaries. The hidden representation distributions of phonemes and natural classes in all four conditions were evaluated.

In addition to the scatter plots, we conducted statistical analyses to evaluate the model's capability in discriminating between various phoneme pairs and natural classes. We performed a multivariate statistical test employing Hotelling's T-square test. The statistical power of the test is inevitably affected by the sample size and when testing the entire set of data points, every pair under comparison was considered significantly different. In response to this issue, we implemented multiple tests with randomly selected 25 samples for each group. In order to alleviate the potential biases arising from sampling, we executed 500 random samplings and tests for each pair. Subsequently, we computed hit rate as the percentage of trials showing significant difference among the 500 trials.

Also, the silhouette score between the two phonemes or natural classes was computed. The silhouette score quantifies the degree of overlap and separability between distinct clusters. It is a metric that falls within the range of -1 to 1, where 0 signifies overlapping clusters. A score below 0 suggests that many data points are labelled as part of a cluster, which the algorithm believes should belong to another. Conversely, a score above 0 indicates that the two clusters are relatively distinct, with values closer to 1 indicating a higher degree of separation.

To preview, our results demonstrate that in the learned hidden representation space, not only the instances of phones are fairly clustered around categorical centers, but the phone clusters are also organized according to phonological features. Furthermore, despite the typological unrelatedness and divergent distributions of phonological features between Mandarin and English, similar performances were achieved, in line with the ABX test results in Section 3.2.2. It is also observed that when encoding a foreign and typologically different language, the model exhibited comparable patterns, albeit with variations, as with the categorical ABX test results. The model thus appears to rely more on universal properties of sounds to learn distinctive features. In the following sections from 3.3.3 to 3.3.6, we present only representative features and natural classes. A full set of analysis can be found in the link provided in the Google drive link (p. 11).

### 3.3.3  *Vowels*

The results of the phoneme evaluation showed that vowel contrasts were well distinguished, as the example of backness contrast shows in Figure 10 and Table 3. On Mandarin data, the distinction between classes is very clear, with a relatively high degree of separation. On English data, while less distinct, the division was still apparent.

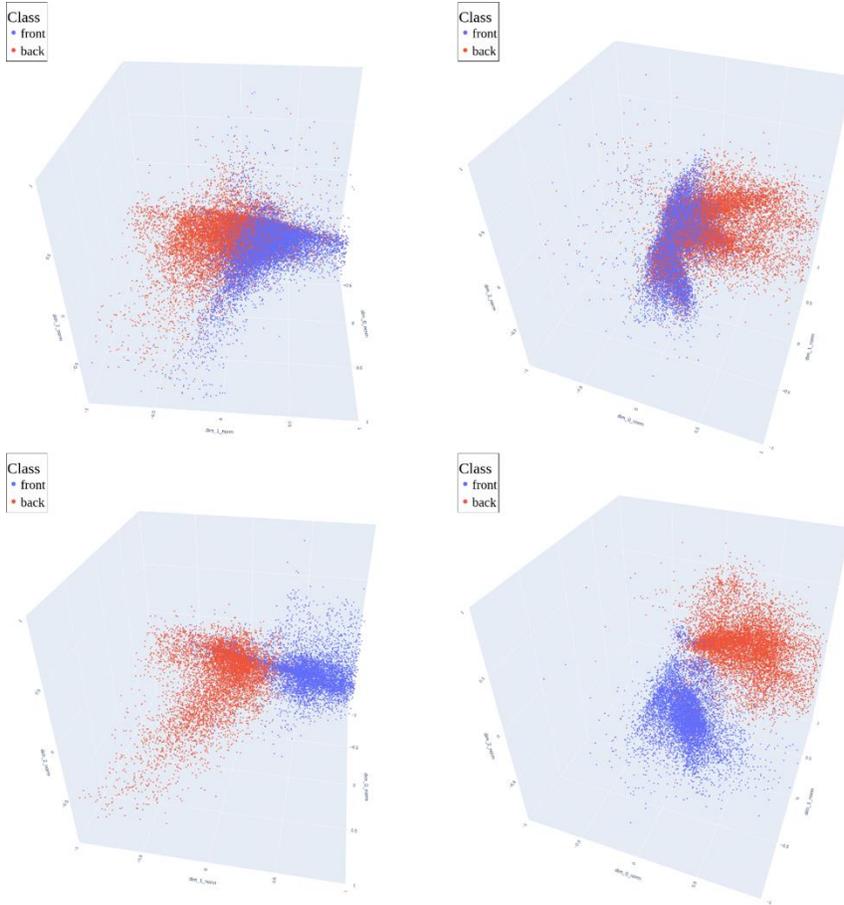

**Figure 10**. Plot of [±back] hidden representations. Upper left is EE, upper right ME, lower left EM and lower right MM. Mandarin natural class members include /i/, /y/, /eɪ/~/o/, /u/, /ɤ/, /ou/; English members include /æ/, /ɛ/, /eɪ/, /ɪ/, /i/~/ɔ/, /ɔɪ/, /oʊ/, /ʌ/, /u/, /ʊ/.

| Indicator | Silhouette | | Hit Rate | |
|---:|:---:|:---:|:---:|:---:|
| *Training Evaluation* | *English* | *Mandarin* | *English* | *Mandarin* |
| *English* | 0.14 | 0.11 | 0.98 | 0.96 |
| *Mandarin* | 0.43 | 0.46 | 1.00 | 1.00 |

**Table 3** Silhouette score and hit rates between [±back] pairs in the four conditions.

In Table 3, it could be observed that a rather clear distinction of [+back] and [-back] exists in all four conditions: the shape and direction of the potential decision boundary remain consistent. This observation is consistent with the ABX testing results. Regardless of the language being learned, the model demonstrates sensitivity to other languages similar to how early infant language learners would respond.

In contrast to the relatively clear distinctions observed for most features associated with vowels, the acquisition of [±diphthong] was not successful, as shown in Table 4 and Figure 11. The boundaries between monophthongs and diphthongs are indistinct, resulting in a significant overlap between the two regions.

| Indicator | Silhouette | | Hit Rate | |
| --- | --- | --- | --- | --- |
| *Training / Evaluation* | *English* | *Mandarin* | *English* | *Mandarin* |
| *English* | 0.03 | 0.003 | 0.18 | 0.08 |
| *Mandarin* | 0.05 | 0.01 | 0.55 | 0.38 |

**Table 4** Silhouette score and hit rates between [±diphthong] pairs in the four conditions.

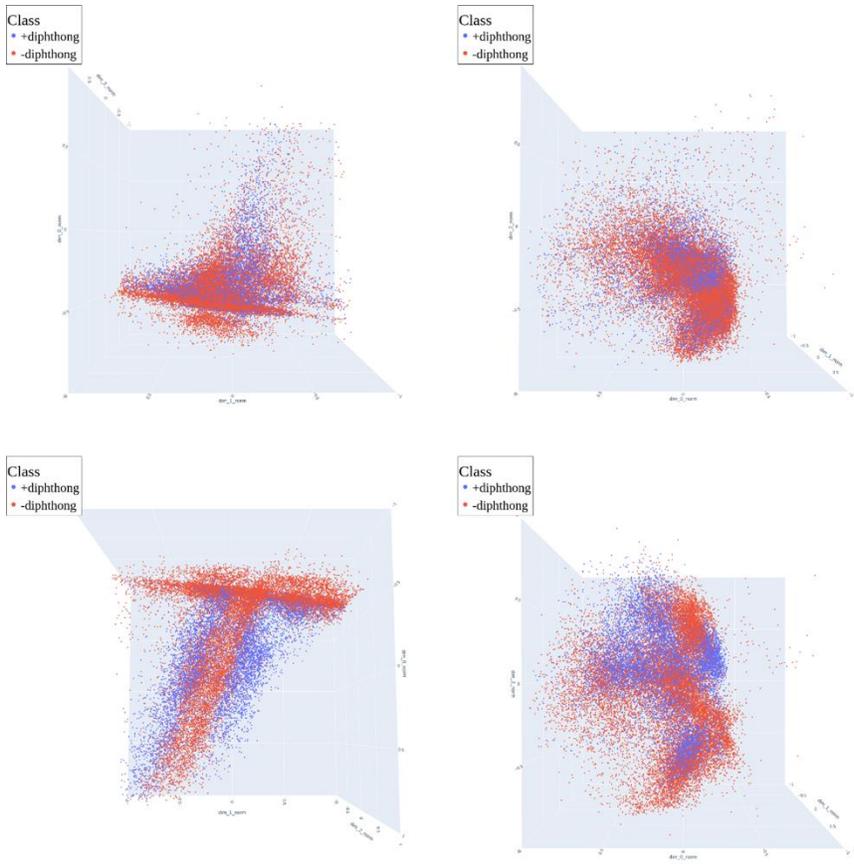

**Figure 11**. Plot of [±diphthong] hidden representations. English natural class members include /aʊ/, /ɔɪ/, /oʊ/, /eɪ/~/ɔ/, /ɛ/, /ɚ/, /ʌ/, /u/, /ʊ/, /ɪ/, /i/; Mandarin members include /ou/, /ai/, /eɪ/, /au/~/i/, /ɚ/, /ɿ/, /y/, /u/, /o/, /a/, /ɤ/.

### 3.3.4   *Consonants*

The patterns of consonants are generally more complex and varied than those of vowels. However, the models have successfully learned many of the consonant pairs, see Figure 12 – 15 and Table 5 – 9. These plots illustrate the model's ability to encode contrasts associated with consonants into its hidden space, thereby facilitating decoding. In this structured representation, not only are more dissimilar sounds positioned farther apart, but they also inhabit distinct locations within the hidden space. This organization exhibits a degree of alignment with established phonological features. Details are as follows.

An example of successfully acquired features is the [±strident] distinction. The results are shown in Figure 12 and Table 5. Despite a higher frequency of affricates in Mandarin compared to English, the English learner model did not exhibit a significant disadvantage in learning affricates compared to Mandarin, especially when evaluating both on Mandarin data, suggesting a potential cross-linguistic universality in learning of the model to some extent.

| Indicator | Silhouette | | Hit Rate | |
| --- | --- | --- | --- | --- |
| *Training \ Evaluation* | *English* | *Mandarin* | *English* | *Mandarin* |
| *English* | 0.19 | 0.15 | 0.99 | 1.00 |
| *Mandarin* | 0.16 | 0.17 | 1.00 | 1.00 |

**Table 5** Silhouette score and hit rates between [±strident] pairs in the four conditions.

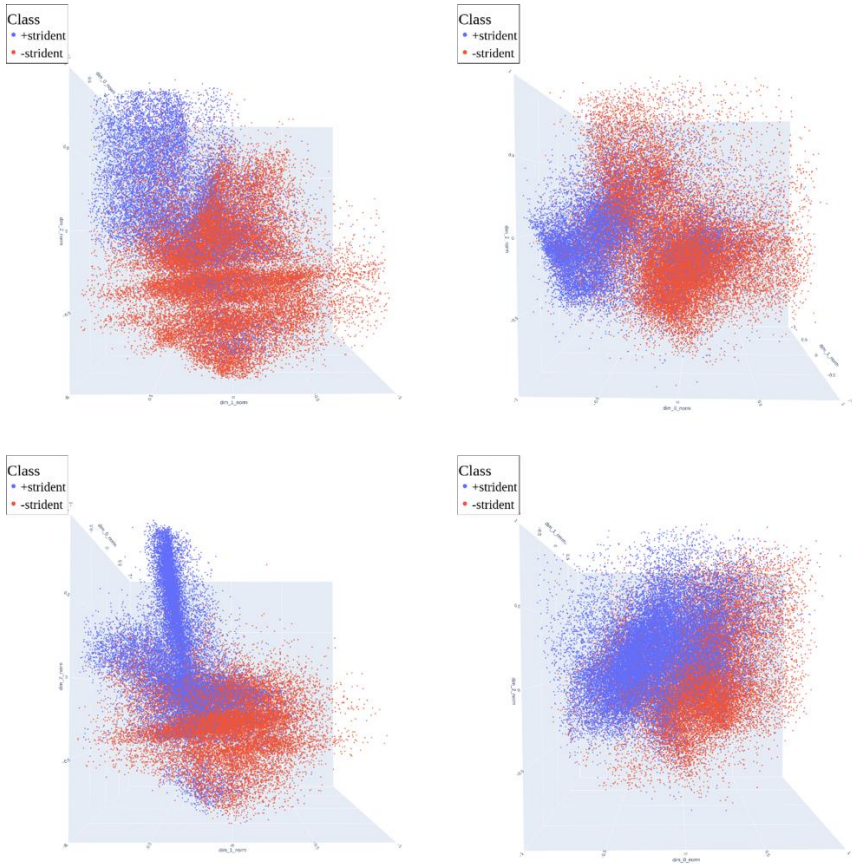

**Figure 12**. Plot of [±strident] hidden representations. English natural class members include /s/, /z/, /ʃ/, /ʒ/, /tʃ/, /dʒ/, /f/, /v/~/t/, /g/, /k/, /d/, /b/, /p/, /h/, /m/, /n/, /l/, /ɹ/, /j/, /w/; Mandarin members include /ts/, /tsʰ/, /s/, /tɕ/, /tɕʰ/, /ɕ/, /tʂ/, /tʂʰ/, /ʂ/, /ʐ/, /f/~/tʰ/, /k/, /kʰ/, /t/, /p/, /pʰ/, /x/, /m/, /n/, /l/.

Recall that there are significant differences in the use of voicing contrast between the two languages (see Section 2.1). Despite this distinction, our modeling results in Figure 13 and Table 6 demonstrate that both models successfully acquired knowledge of the voicing contrast in all conditions: although Mandarin has only one voicing contrast pair (/ʂ/~/ʐ/), in contrast to its high frequency in English, the model trained on Mandarin data exhibits comparable performance when evaluated on English, akin to the results obtained by the model trained on English data. Similar tendency was also observed in ABX test in Section 3.2.2.

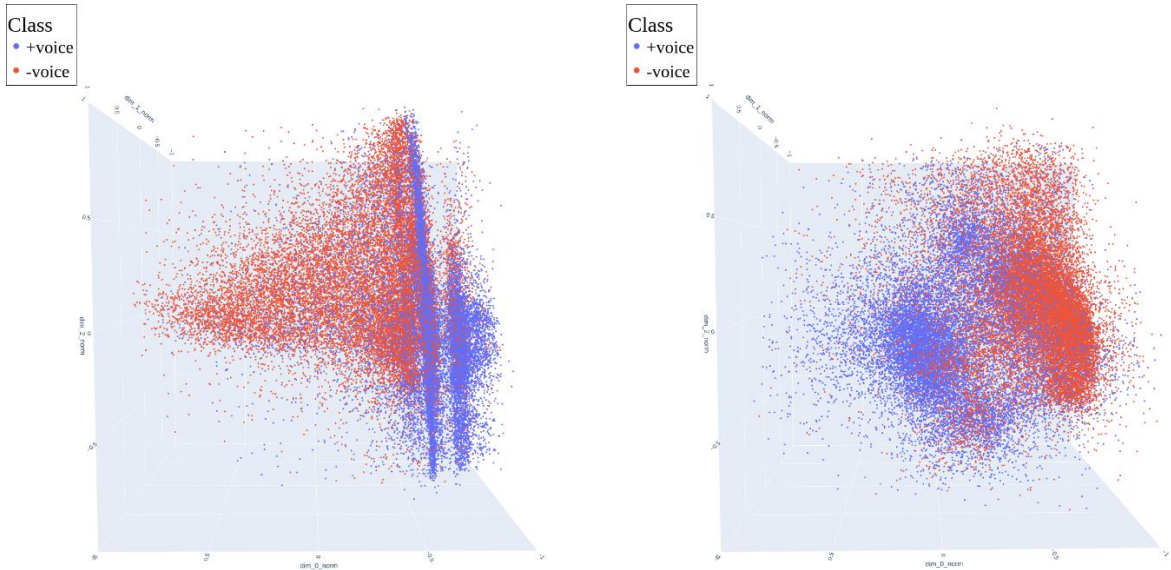

**Figure 13**. Plot of [±voiced] hidden representations evaluated on English dataset (left EE, right ME). Members include /d/, /dʒ/, /ð/, /ʒ/, /z/, /ɡ/, /b/, /v/~/t/, /tʃ/, /θ/, /ʃ/, /s/, /k/, /p/, /f/.

| Indicator | Silhouette | | Hit Rate | |
|---|---|---|---|---|
| *Training Evaluation* | *English* | *Mandarin* | *English* | *Mandarin* |
| *English* | 0.11 | 0.11 | 0.96 | 0.94 |

**Table 6** Silhouette score and hit rates between [±voiced] pairs in the four conditions.

A few features associated with consonants were not acquired successfully. For example, our evaluation of the nasal feature [±nasal] indicates that the distinction was not salient in either Mandarin or English, with considerable overlap in the class distributions. See the results in Figure 14 and Table 7.

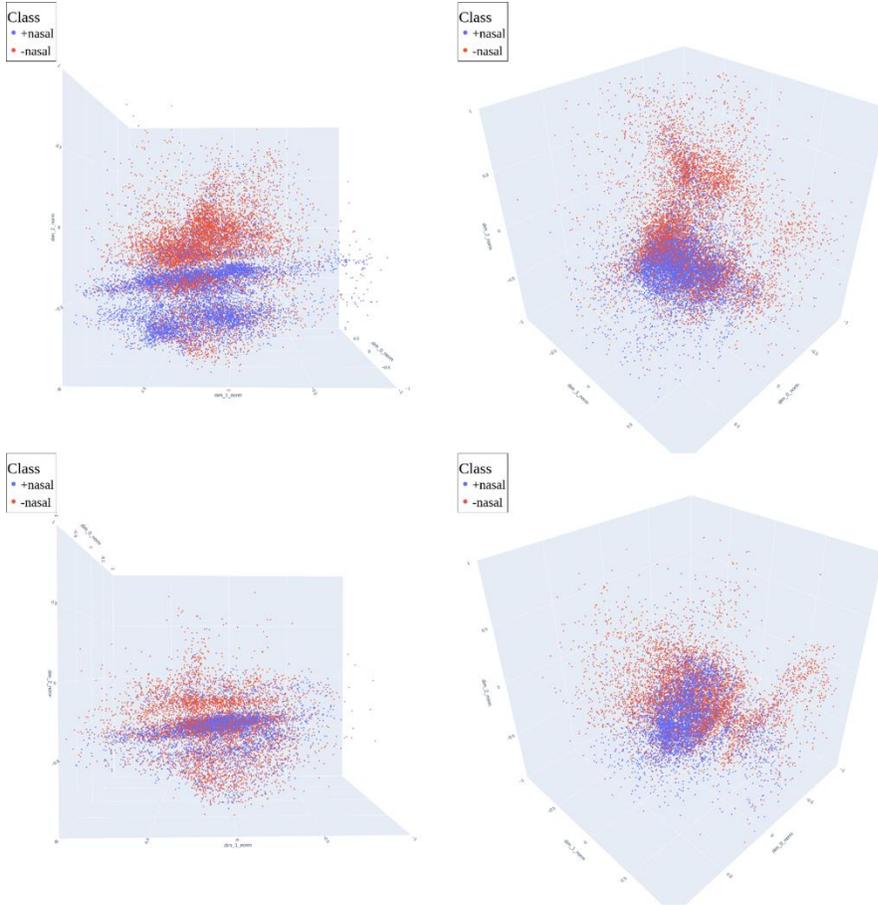

**Figure 14**. Plot of [±nasal] hidden representations. English natural class members include /n/, /m/, /ŋ/~/d/, /b/, /g/; Mandarin members include /n/, /m/~/t/, /p/.

| **Indicator** | **Silhouette** | | **Hit Rate** | |
|---|---|---|---|---|
| *Training \ Evaluation* | *English* | *Mandarin* | *English* | *Mandarin* |
| *English* | 0.09 | 0.08 | 0.90 | 0.77 |
| *Mandarin* | 0.01 | 0.04 | 0.26 | 0.49 |

**Table 7** Silhouette score and hit rates between [±nasal] pairs in the four conditions.

As shown from the descriptive data above (Figure 14 and Table 7), both the nasality contrast and place distinctions within nasal sounds seem hard to the model. In both Mandarin and English, either evaluated with exposed or foreign language, nasal sounds at different places of articulation were projected very close to each other with great overlap (see Table 8). Especially, the distinction between /n/ and /ŋ/ was extremely small in both languages. This bad performance seems to conform with human infants' difficulty in distinguishing these two nasal consonants (Narayan et al., 2010). In addition, the distinction between nasal stops and oral stops was small as well (Table 9).

| Silhouette | EE | ME | MM | EM |
|---|---|---|---|---|
| m~n | 0.04 | 0.03 | 0.03 | 0.06 |
| m~ŋ | 0.06 | 0.05 | | |
| n~ŋ | 0.03 | 0.01 | | |

**Table 8** Pairwise silhouette scores of nasal consonants in the four conditions.

| Silhouette | EE | ME | | MM | EM |
|---|---|---|---|---|---|
| b~m | 0.15 | 0.12 | p~m | 0.05 | 0.04 |
| d~n | 0.10 | 0.11 | t~n | 0.05 | 0.01 |
| g~ŋ | 0.05 | 0.06 | | | |

**Table 9** Pairwise silhouette scores between nasal and oral stops in the four conditions. For a full data set, see the Google drive link (p.11).

### 3.3.5   *Feature distance*

The hidden space separateness appears to be consistent with the theory-driven feature-based distance metric, where the similarity between phonemes *a* and *b* is greater than that between *a* and *c* if *a* and *b* have fewer contrastive features than *a* and *c* (Bailey & Hahn, 2005). For the purpose of presentation, we show four phonemes from each of the two languages, in English /t, d, b, m/ and in Mandarin /tʰ, t, p, m/. These phonemes form a (near-)minimal pair continuum where two adjacent phonemes differ in one feature that is contrastive in the language. Specifically, for the English set, the difference between /t/ and /d/ is only in voicing; that between /d/ and /g/ is only in place of articulation ([±labial]); whereas between /b/ and /m/ only nasality. Similarly, the difference between /tʰ/ and /t/ is only in aspiration; /t/ and /p/ differ only in place of articulation; and /p/ and /m/ differ in nasality and voicing, which is not contrastive in Mandarin.

As shown in Table 10 and Table 11, the silhouette scores between pairs of different featural contrasts seem accumulative, although not linear. This might suggest that feature-based distance metric is aligned with empirical acoustic-based quantitative prediction, as in the current study, to the evaluation of phonemic separateness. For example, the two phonemes /t/ and /d/ in English and /tʰ/ and /t/ in Mandarin, which differed only by voicing or aspiration, were closer than /t/ and /b/ or /tʰ/ and /p/, which differed by both voicing / aspiration as well as place of articulation. And these two pairs were closer than /t/ and /m/ or /tʰ/ and /m/, which differed by even more features (English:

$d(\text{t}, \text{d}) < d(\text{t}, \text{b}) < d(\text{t}, \text{m})$; Mandarin: $d(\text{t}^h, \text{t}) < d(\text{t}^h, \text{p}) < d(\text{t}^h, \text{m})$). When the four language conditions are compared, it appears that the ability to learn the relation is consistent across different languages and is applicable to both exposed and foreign language. Yet, by comparing /t/ vs /d/, /d/ vs /b/, and /b/ vs /m/ in English, as well as /tʰ/ vs /t/, /t/ vs /p/, and /p/ vs /m/ in Mandarin, it revealed that the separateness created by various contrastive features may not be uniform. Due to a possibly high dimensionality of hidden representation and the non-uniform featural distances, the separateness created by aspiration and that by place of articulation may not be directly comparable, i.e., it is not guaranteed that $d(\text{t}^h, \text{t}) > d(\text{t}^h, \text{p}^h)$, for example.

| Silhouette | MM | | | EM | | |
|---|---|---|---|---|---|---|
| | t | p | m | t | p | m |
| tʰ | 0.11 | 0.12 | 0.25 | 0.12 | 0.12 | 0.23 |
| t | | 0.01 | 0.08 | | 0.02 | 0.07 |
| p | | | 0.06 | | | 0.04 |

Table 10. Pairwise silhouette scores of /tʰ, t, p, m/ for MM and EM.

| Silhouette | EE | | | ME | | |
|---|---|---|---|---|---|---|
| | d | b | m | d | b | m |
| t | 0.13 | 0.14 | 0.30 | 0.10 | 0.11 | 0.27 |
| d | | 0.07 | 0.14 | | 0.03 | 0.07 |
| b | | | 0.15 | | | 0.12 |

Table 11 Pairwise silhouette scores of /tʰ, t, p, m/ for EE and ME.

### 3.3.6    *Allophonic distribution*

Now we move on to the learning of allophonic distinctions. As an illustration, we present the allophonic distributions of /i, ɨ, ɻ/ in Mandarin (Figure 15 and Table 12). This group was chosen for presentation purposes here, due to its comparable number of tokens in the Mandarin dataset. Our objective was to investigate potential differences in the learning of allophonic distributions compared to phonemic distinctions.

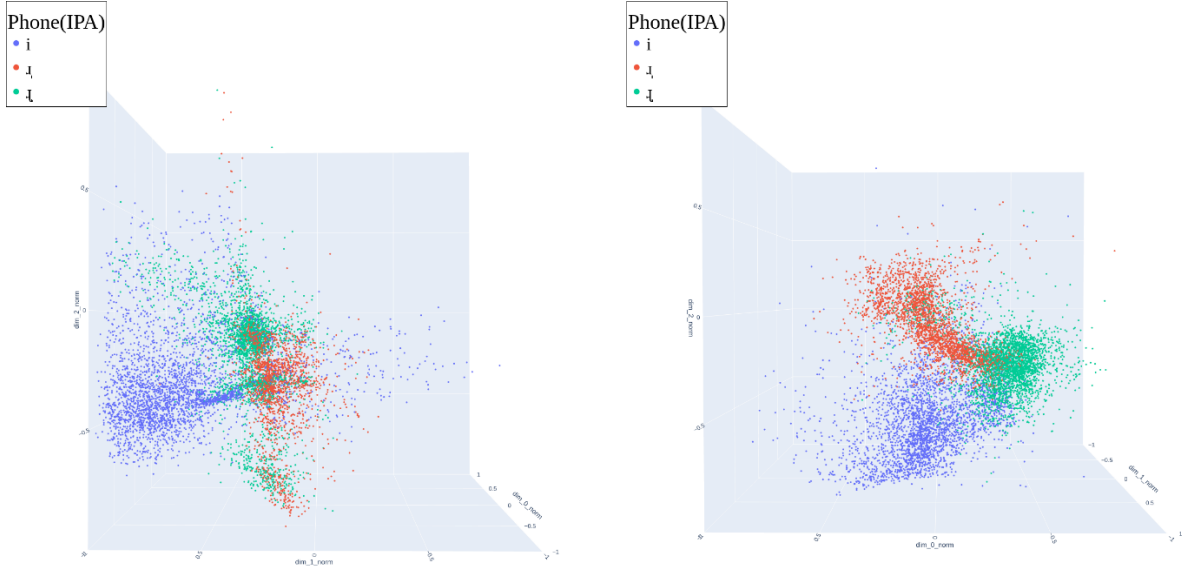

**Figure 15**. Plot of variants of *i* hidden representations evaluated on Mandarin dataset (left EM, right MM). Members include i~/i/, ii~/ɨ/, iii~/ɯ/.

| Silhouette | MM | EM |
|---|---|---|
| i~ɨ | 0.44 | 0.33 |
| i~ɯ | 0.45 | 0.27 |
| ɨ~ɯ | 0.30 | 0.12 |

**Table 12** Pairwise silhouette scores between the three allophones of /i/ in MM and EM.

The three allophones were well distinguished as in Figure 15 and Table 12, but importantly the current bottom-up models did not show a sensitivity to the difference between phoneme and allophone projections. When tested on ABX test, the pairwise error rate among vowel phonemes are not significantly different from the pairwise error rate among allophones ($\mu_V = 0.15, \sigma_V = 0.146; \mu_A = 0.182, \sigma_A = 0.084; p = 0.371$); neither one is more distinct (as in Kolanchina & Magyar, 2019) or less distinct. In a sense, allophony was not captured by the model. Unlike top-down models (Kolanchina & Magyar, 2019; Silfverberg et al., 2018), and despite human's different perceptual sensitivity and learnability of phonemes and allophones (Martin et al., 2013, Mitterer et al., 2018, Pepperkamp et al., 2003), the current model was not able to tell whether a contrast is phonemic or allophonic. This failure in learning allophonic distributions is expected because the model had limited exposure to contextual information. Without sufficient exposure to phonological context, it becomes challenging for the model to differentiate between allophones and phonemes, as shown from human learners (Maye & Gerken, 2001; Pepperkamp et al., 2003).

## 4 General discussion

By training an autoencoder model on bottom-up data, this study explored the extent to which context-free acoustic information can bring learners to in terms of phonetic acquisition. To do so, we used evaluation methods including clustering test, ABX test, as well as direct visualization of hidden representation space. The utilization of unsegmented, non-transcribed "raw" sound data aligns with the initial challenges encountered by human infants during the very early stages of language acquisition, where they are exposed to continuous, unsegmented sound streams as input without much access to top-down linguistic information (Liberman et al., 1974). The consistent results across various evaluation methods suggest that clusters resembling phonetic categories can emerge from training on acoustic information alone. The cross-linguistic consistence further indicates a degree of universality, regardless of the statistical availability of contrasts in the exposed language. Additionally, the model's insensitivity to the distinction between phonemes and allophones indicates a lack of phonological knowledge.

### 4.1 Emergence of prototypical categories

The current experimental results support the hypothesis that knowledge about prototypical categories resembling phonetic categories can be learned through "repeating". Specifically, this process involves iteratively projecting sounds into a hidden space, and subsequently reconstructing this space into a production without prior knowledge of segmentation or phoneme categories. The acquired knowledge is represented by different distributions of phones in the hidden space. The model was capable of clustering instances of the same phonemes and projecting different phonemes to separate regions in the hidden space. The projection further appeared to group similar sounds sharing features together in closer regions and separate dissimilar sounds in farther regions. While the model was not perfect, it demonstrated successful acquisition as evidenced by significantly higher HCV scores compared to random chance in the clustering evaluation.

The knowledge acquired by the current model can be likened analogous to Kuhl's (1991) phonetic category centers. In the assumed mental representation space where "category members" (analogous to hidden representations of segments) were located, instead of randomly dispersed, the representations of the input segments were clustered into categories that align with phone categories, and the categories are further distributed along dimensions reflecting the featural contrast.

Recall that the learning process in the current study is bottom-up: the model is primarily trained with acoustic data while minimizing the incorporation of distributional information. During the model's acquisition phase, it concentrates on the sounds themselves. This parallels the very early stages of human language acquisition, where infants, although lacking an understanding of the language, immerse themselves in linguistic input. During this stage, learners gradually develop distinct phonetic protocategories, primarily based on the physical properties of the sounds (Grieser & Kuhl, 1989; Kuhl, 1991; Samuel, 1982).

## 4.2 Cross-linguistic similarities and discrepancies between the feature availability and their learned outcomes

Our experiment with the general ABX test found that the model in general showed a preference for the exposed language, displayed by a lower ABX error rate compared to tests in a foreign language. However, when tested on specific contrasts that are minimally contrastive in terms of phonological features, the model did not show a significant preference for the exposed language. These results align with previous experiments on infants aged 6–8 months, where both native and non-native infants demonstrated similar discrimination abilities between sounds. The tested contrasts were chosen to represent the different distributions in the two languages, aiming to provide a comprehensive understanding of the relationship between the availability of certain contrasts in the input language and the learners' discrimination abilities. We selected contrasts that are existent in one language but not in the other, contrasts that are prevalent in one language but rare in the other, and contrasts that are common in both languages. Despite these varying distributions, the tested contrasts showed no significant performance differences across the two languages. This suggests that after training on purely bottom-up data, the model did not modify their mental representations in favor of the exposed language, thereby sacrificing the ability to discriminate non-native contrasts. However, the fact that the model's ability to remain relatively universal and sensitive to contrasts non-native to the exposed language does not imply that it is not learning from the input.

As shown in these tests, features are learned with varying degrees of success within each language. This variation does not strictly reflect the availability of these features in the input language as well. Notably, certain features were learned equally well, despite their different representations in Mandarin and English. For instance, both the Mandarin and English models showed a remarkable ability to acquire voicing contrast, despite clear differences in the number of consonantal pairs exhibiting this feature in the two languages. One potential factor to consider is the acoustic salience of well-acquired features. Consider voicing as an example. In terms of acoustic cues, voicing is clearly represented by a periodic soundwave and continuous fundamental frequency (F0), whereas voiceless sounds lack this characteristic feature (Johnson, 2011, p.13). This prominent contrast between voiced and voiceless sounds could have potentially facilitated the learning of voicing feature, even in a language where the contrast was very limited. Another relevant factor that may contribute to the exceptional learning of voicing contrast in Mandarin is the structure of the learner model. The model here employed a vector-based representation that flattened time slices into vectors, rather than considering time series sequentially. As a result, the model may have effectively learned to recognize instantaneous and non-sequential phonetic indicators, rather than sequential ones. Since the voicing feature is present continuously and consistently throughout an entire segment, the model could detect it upon encountering even a small slice with voicing. And from both our experimental assumptions and empirical evidence, it is clear that the model does not develop representations based on phonological contrast or adjust its representations according to the statistical distributions of sounds in the language. Consequently, despite the relative rarity of voicing contrast in Mandarin, the model is remarkably proficient at learning voicing contrast.

In contrast, although aspiration is widely contrastive in Mandarin, with all plosives and affricates exhibiting minimal pairs that are contrastive in aspiration, the model's learning outcome was unsatisfactory ($S = 0.04$). If the previously mentioned hypothesis regarding sequential indicators holds true, this outcome could be attributed to the fact that detecting aspiration necessitates the model's processing of sequential elements, such as identifying a voiceless burst immediately followed by silence, which proved to be a challenging task for the model to learn.

Another feature that was poorly acquired in both language models was nasality, which exhibits salient, continuous, and consistent acoustic cues (Johnson, 2011, pp.185-191). In fact, it is not unprecedented for nasality to be a challenging aspect to model in phonological learning. For instance, Chen & Elsner (2023) reported the difficulty of learning nasality, though their focus was on nasality in vowels shown in phonological contexts. One potential explanation for this observation is that nasal consonants spread their nasal features along the temporal sequence (Thompson, 1978). As a result, sounds preceding and following a nasal consonant tend to exhibit nasal features as well. Therefore, the model may have incorrectly learned to attribute this feature to a broad range of sounds, thus across multiple data frames, leading to a false overgeneralization of the nasal feature. As the current paper does not specifically investigate the learnability of individual features and contrasts and their underlying mechanisms, we did not conduct hypothesis testing in this regard. However, we acknowledge that this topic requires further research in the future.

## 4.3 The role of top-down phonological learning

It is important to note that the strict differentiation between bottom-up and top-down learning does not necessarily correspond to distinct developmental stages in speech acquisition. In reality, infants have access to both phonetic and distributional cues of sounds from the beginning. However, the current experiment exclusively analyzed the performance of a model that underwent bottom-up learning only, without using distributional information. The purpose of the current exercise was to accurately assess the scope of learning achieved through a pure bottom-up approach, rather than to simulate the precise learning processes that infants undergo. The outcomes demonstrate the emergence of prototypical phonetic categories, where sound tokens belonging to the same phonetic category is more likely to be clustered together, similar to the protocategories suggested in previous works (Grieser & Kuhl, 1989; Kuhl, 1991; Samuel, 1982).

Without top-down information, the clusters in the model resemble phonetic categories, rather than phonemic. For instance, the three allophones of Mandarin "i" (/i, ɿ, ʅ/) (Zee & Lee, 2001) were projected to separate regions in the hidden space and was not significantly different from other phonemic vowels. The model's inability to cluster allophones more closely together than phonemes may indicate one boundary between the contributions of bottom-up and top-down information. Learners acquire a fundamental, natural, cross-linguistic hidden representation space from bottom-up information, but it is essential to use top-down information to transform this space to reflect paradigmatic relationships and co-occurrence restrictions (Kolachina & Magyar, 2019).

## 5 Conclusion and limitations

To conclude, the present study has demonstrated that an autoencoder model, from training on unsegmented and non-transcribed sound data, not only captures the distinction between phones and forms phonetic protocategories, but also learns to map these phonetic protocategories into the hidden representation space based on phonological features. Through cross-linguistic comparisons and foreign language evaluations, it is found that the model acquires knowledge reflecting the fundamental, universal properties of sounds across languages, while it does not acquire language-specific phonological knowledge about paradigmatic relationships and co-occurrence restrictions. It displays the extent that context-free acoustic information as input can bring learners to in early phonetic acquisition, which is about the first half of year after birth.

It must be acknowledged that there are limitations to our model. We deliberately employed basic models in an effort to minimize the influence of top-down information. However, the simple autoencoder model that relies on fully-connected layers lack the ability to process information sequentially, which is inherently essential for handling sound streams. The limitation may not be as critical since the model was fed with very small randomly segmented recordings. Nevertheless, the resampling technique, although applied in previous works (Kamper et al., 2015; Shain & Elsner, 2019), may have altered various acoustic cues, potentially resulting in less extended features for longer sounds compared to shorter ones. To effectively model more sequential or distributional phonological knowledge, such as phonotactic patterns that correspond to later stages of phonological acquisition, further modifications will be necessary for our current model.

Compared to other existing models like GANs (Beguš, 2020b), our current model's architecture is not as biomimetic because its training objective is to precisely reconstruct the input, or "repeating" exactly what it heard. However, at the very early stage of phonological acquisition, which was the focus of our study, articulations have only just started (Kuhl, 2004), and feedback-based adjustment of phonetic and phonological knowledge may not be critically influencing learners' acquisition. Therefore, especially for the simulation of early phonetic learning, the model architecture of an autoencoder model may still provide meaningful insights into phonological acquisition. Future research that simulates the later stages of phonological acquisition may benefit from applying more justifiable model structures or different models.

In addition, it remains to be explored the extent to which poorly-learned features and well-learned features reflect the natural properties of sounds that make them difficult to distinguish and how exactly the model's availability to distinguish different features is compatible with that of human learners at the early stages of phonological acquisition. Given that we have not specifically tested the exact properties in deriving these results, we would like to refrain from further speculation. However, the investigation of which properties of acoustic input lead to universal learning outcomes is an area that warrants further exploration in the modeling literature.

Overall, by exploring how far purely bottom-up input can take a learner in early phonetic learning and demonstrating actual shapes and structures of the hidden space of phones and phonological features, this study extends from previous research. By doing so, it contributes to our understanding of the underlying mechanisms involved in the acquisition of phones and phonological features during earlier stages. Further studies using other models and more diverse language datasets could provide additional insights into the early stage acquisition of phonetic and phonological knowledge.